\documentclass[10pt,twocolumn,letterpaper]{article}

\usepackage{wacv}
\usepackage{times}
\usepackage{epsfig}
\usepackage{graphicx}
\usepackage{amsmath}
\usepackage{amssymb}
\usepackage{caption}
\usepackage{multirow}
\usepackage[bookmarks=false]{hyperref}
\usepackage{float}

\wacvfinalcopy 

\def\wacvPaperID{993} 

\ifwacvfinal\fi
\begin{document}

\newcommand\blfootnote[1]{%
  \begingroup
  \renewcommand\thefootnote{}\footnote{#1}%
  \addtocounter{footnote}{-1}%
  \endgroup
}

\title{AlignNet: A Unifying Approach to Audio-Visual Alignment}

\author{Jianren Wang*\\
CMU\\
{\tt\small jianrenw@andrew.cmu.edu}
\and
Zhaoyuan Fang*\\
University of Notre Dame\\
{\tt\small zfang@nd.edu}
\and
Hang Zhao\\
MIT\\
{\tt\small hangzhao@csail.mit.edu}
}

\twocolumn[{
\maketitle
\ifwacvfinal\thispagestyle{empty}\fi
\begin{center}
\includegraphics[width=\textwidth]{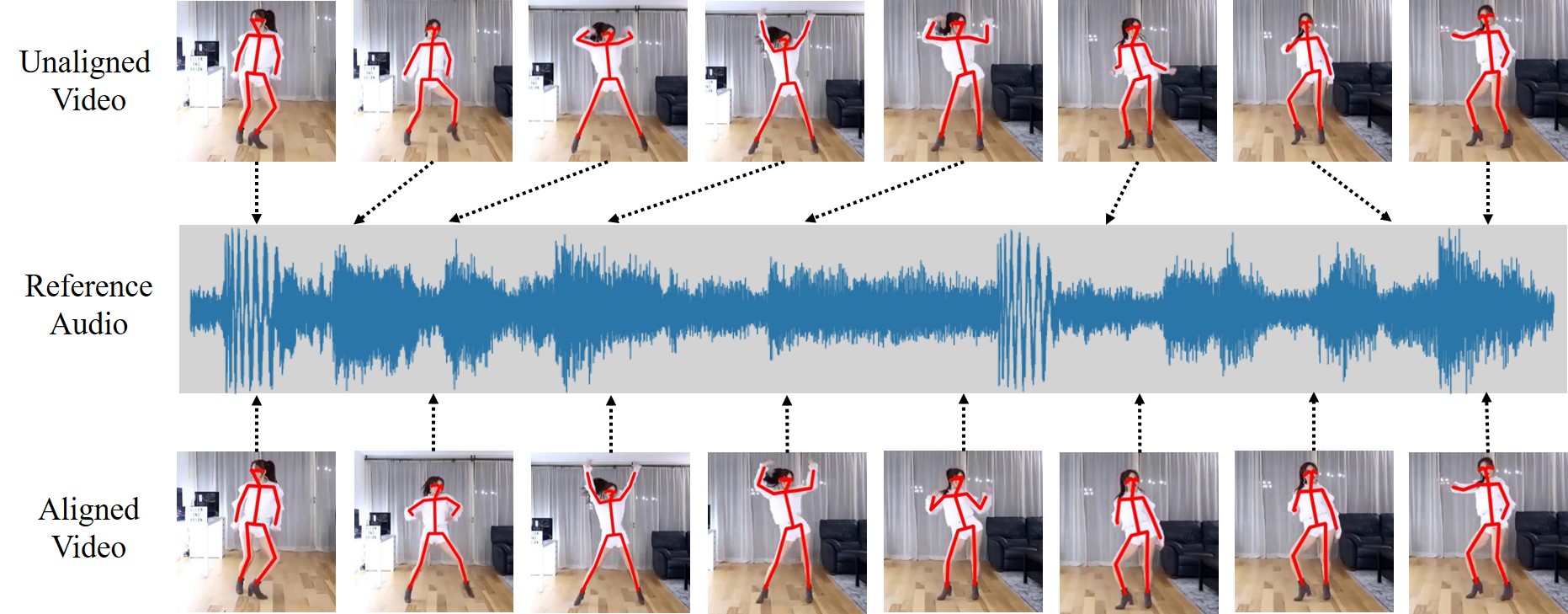}\\
\captionof{figure}{Given a pair of unaligned audio and video, our model aligns the video to the audio according to the predicted dense correspondence.}
\label{fig:teaser}
\end{center}
}]

\blfootnote{* indicates equal contribution}

\begin{abstract}
We present AlignNet, a model that synchronizes videos with reference audios under non-uniform and irregular misalignments. AlignNet learns the end-to-end dense correspondence between each frame of a video and an audio. Our method is designed according to simple and well-established principles: attention, pyramidal processing, warping, and affinity function. Together with the model, we release a dancing dataset \textit{Dance50} for training and evaluation. Qualitative, quantitative and subjective evaluation results on dance-music alignment and speech-lip alignment demonstrate that our method far outperforms the state-of-the-art methods. Code, dataset and sample videos are available at our project page\footnote{\url{https://jianrenw.github.io/AlignNet/}}.

\end{abstract}


\section{Introduction}
Dancers move their bodies with music, speakers talk with lip motions and are often accompanied by hand and arm gestures. The synchrony between visual dynamics and audio rhythms poses perfect performances. However, recorded videos and audios are not always temporally aligned. \textit{e.g.} dancers may not be experienced enough to follow the music beats precisely; ``Automated Dialogue Replacement (ADR)" is used in film making instead of simultaneous sounds for lip synchronization. It is not hard to imagine the humongous amount of efforts and time required to temporally synchronize videos and audios by aligning visual dynamics and audio rhythms. This alignment problem is especially difficult for humans since the misalignment is often non-uniform and irregular.

There are several previous attempts to address this problem.
\cite{davis2018visual} extracted visual beats analogous to musical beats, and applied audio-video alignment to tasks such as video dancification and dance retargeting. Their method requires feature engineering on visual beats and only performs alignment on beat-level, which makes the generated dances sometimes exaggerated.
SyncNet~\cite{chung2016out} proposed to learn the alignment features with a contrastive loss that discriminates matching pairs from non-matching pairs. However, they assume a global temporal offset between the audio and video clips when performing alignment.
\cite{halperin2019dynamic} further leveraged the pre-trained  visual-audio features of SyncNet~\cite{chung2016out} to find an optimal alignment using dynamic time warping (DTW) to assemble a new, temporally aligned speech video. Their method can therefore stretch and compress the signals dynamically. However, the minimum temporal resolution of their method is limited (0.2 second).
In a realistic setting, \textit{e.g.}~ADR or off-beat dancing, misalignment between audio and video can happen at any moment on arbitrary temporal scale. Therefore, a crucial property for audio-visual alignment solutions is the ability to deal with arbitrary temporal distortions. Another nice property to have is end-to-end training, as it significantly simplifies the solution to such a hard task.



In this work, we present AlignNet, an end-to-end trainable model that learns the mapping between visual dynamics and audio rhythms, without the need of any hand-crafted features or post-processing. Our method is designed according to simple and well-established principles: attention, pyramidal processing, warping, and affinity function. First, attention modules highlight the important spatial and temporal regions in the input. Then, casting in two learnable feature pyramids (one for video and one for audio), AlignNet uses the current level correspondence estimation to warp the features of the reference modality (video or audio). Warped features and reference signal features are used to construct an affinity map, which is further processed to estimate denser correspondence.

Together with the model, we introduce a dancing dataset~\textit{Dance50} for dance-music alignment. The dataset is cleaned and annotated with human keypoints using a fully automated pipeline.

We demonstrate the effectiveness of AlignNet on both dance-music alignment and lip-speech alignment tasks, with qualitative and quantitative results. Finally we conduct dance and speech retargeting experiments, and show the generalization capabilities of our approach with subjective user studies.

\section{Related Works}

\paragraph{Audio-video alignment}

Audio-video alignment refers to the adjustment of the relative timing between audio and visual tracks of a video. Automatic audio-video alignment has been studied over decades in computer vision.
Early works like \cite{bredin2007audiovisual} and \cite{sargin2007audiovisual} used canonical correlation analysis (CCA) for synchronization prediction.
Later methods tried to align video and audio based on handcrafted features. Lewis~\cite{lewis1991automated} proposed to detect phonemes (short units of speech) and subsequently associate them with mouth positions to synchronize the two modalities. Conversely, \cite{furukawa2016video} classified parameters on the face into visemes (short units of visual speech), and used a viseme-to-phoneme mapping to perform synchronization. \cite{davis2018visual} split videos according to visual beats and applied video-audio alignment on beat-level. 

Learning multimodal features is a recent trend. SyncNet~\cite{chung2016out} learned a joint embedding of visual face sequences and corresponding speech signals in a video by predicting whether a given pair of face sequence and speech track are synchronized.
Similarly, \cite{owens2018audio} proposed a self-supervised method to predict the alignment of motion and sound within a certain time shift. These works attempted to detect and correct a global error, which is a common problem in TV broadcasting. However, they cannot address non-uniform misalignment, \textit{e.g.} dancers do not only make mistakes at musical beats. In other words, misalignment in videos and audios is oftentimes completely unconscious and irregular. In these scenarios, the closest method to our work is proposed by \cite{halperin2019dynamic}, which can stretch and compress small units of unaligned video signal to match audio signal. However, their method can only adjust audio-video misalignment on a coarser granularity, since they assume the consistency of information within every 0.2 second.

\paragraph{Time Warping}
Given two time series, $X = [x_1, x_2, ... , x_{n_x}] \in R^{d\times n_x}$ and $Y = [y_1, y_2, ... , y_{n_y}] \in R^{d\times n_y}$, dynamic time warping (DTW)~\cite{rabiner1993fundamentals} is a technique to optimally align the samples of X and Y such that the following sum-of-squares cost is minimized: $C(P) = \sum_{t=1}^{m} \|x_{p_t^X}-y_{p_t^y}\|^2$, where $m$ is the number of indices (or steps) needed to align both signals. Although the number of possible ways to align $X$ and $Y$ is exponential in $n_x$ and $n_y$, dynamic programming~\cite{bertsekas1995dynamic} offers an efficient way to minimize $C$ using Bellman’s equation. The main limitation of DTW lies in the inherent inability to handle sequences of varying feature dimensionality, which is essential in multimodal data, like video and audio. Furthermore, DTW is prone to failure when one or more sequences are perturbed by arbitrary affine transformations. To this end, the Canonical Time Warping (CTW)~\cite{zhou2009canonical} is proposed, which elegantly combines the least-squares formulations of DTW and  Canonical Correlation Analysis (CCA)~\cite{de2012least}, thus facilitating the utilization of sequences with varying dimensionality, while simultaneously performing feature selection and temporal alignment. Deep Canonical Time Warping (DCTW)~\cite{trigeorgis2016deep} further extends CTW through the usage of deep features. These methods are widely used in audio-audio alignment~\cite{king2012noise,luo2018singing}, video-audio alignment~\cite{halperin2019dynamic} and video-video alignment~\cite{trigeorgis2017deep}.

\paragraph{Audio-visual Datasets}

Most existing audio-visual datasets focus on speech and face motion. Some early datasets are obtained under controlled conditions: forensic data intercepted by police officials~\cite{vloed2014nfi}, 
speech recorded from mobile devices~\cite{woo2006mobile}, \etc. In contrast, VoxCeleb~\cite{nagrani2017voxceleb,Chung2018VoxCeleb2} and AVSpeech~\cite{ephrat2018looking} collected a large amount of audio-visual data of human speeches in the wild.
MUSIC~\cite{Zhao_2018_ECCV,zhao2019sound} and FAIR-Play~\cite{gao20192} are datasets of music and instruments.
In this work, we introduce a new dancing dataset to the community, focusing on the synchronization between music and body motions.
\section{Approach}

\setcounter{figure}{1}
\begin{figure*}
    \centering
    \includegraphics[width=0.9\textwidth]{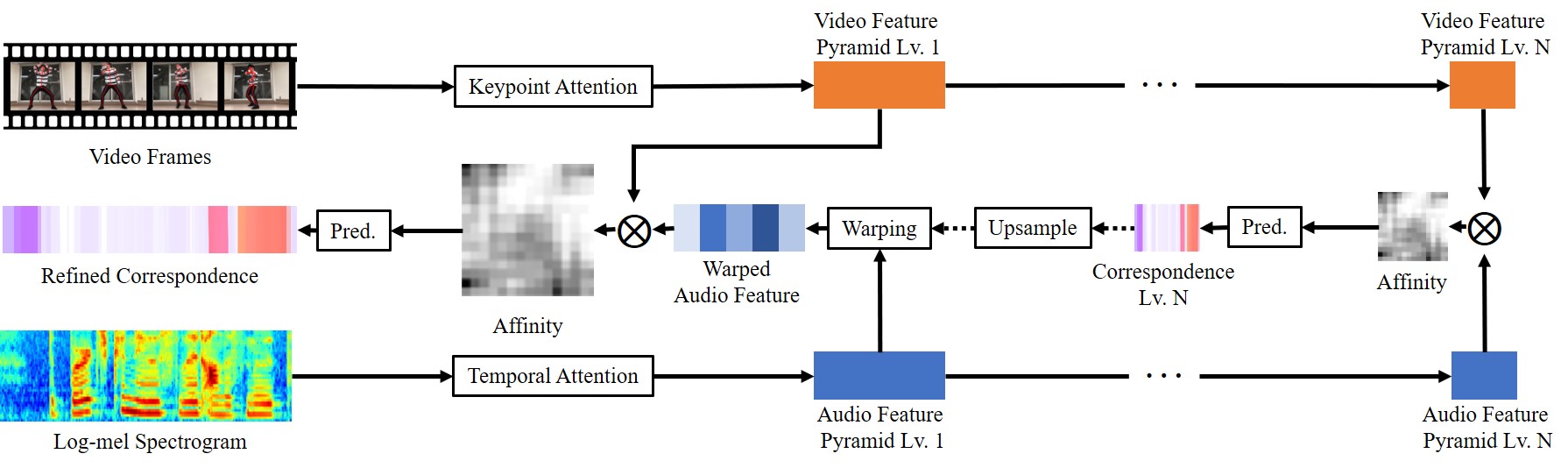}
    \caption{Our model consists of attention, pyramidal processing, time warping, affinity and correspondence prediction.}
    \label{fig:pipeline}
\vspace{-1em}
\end{figure*}

\subsection{Formulation}
AlignNet is an end-to-end trainable model that learns the implicit mapping between visual dynamics and audio rhythms at multiple temporal scales, without the need of any post-processing.
It takes in a video feature sequence $\mathbf{F}_{v}^{t},\: t\in(1,..,n)$ and an audio feature sequence $\mathbf{F}_{a}^{t},\: t\in(1,..,m)$, to predict a dense correspondence $\mathbf{d}$, which is the temporal displacement for each video frame ($n,m$ might not be the same). To generate unaligned training data, we apply random and non-uniform distortion temporally (speed-ups and slow-downs) to the aligned audio-video data of different lengths.

Different from previous methods~\cite{chung2016out,chung2018PerfectMI,halperin2019dynamic} that use raw image frames as inputs, which are prone to large variations when there are changes in outfits, makeups, lighting, shadows, \textit{etc.}, we instead use more robust pose/lip keypoint features provided by a keypoint detector OpenPose~\cite{cao2018openpose}.
Furthermore, as suggested by \cite{chiu2018ActionAgnosticHP}, operating in the velocity space improves the results of human pose forecasting. For our task, at time $t$, we input velocity $\mathbf{v}_{v}\left(t\right)$ and the acceleration $\mathbf{a}_{v}\left(t\right)$ of the keypoints. 
For audio, we use normalized log-mel spectrogram as input. Note that the input video features and audio features do not have the same shape and granularity, but our proposed network learns to extract information from different modalities and align them on the same scale.

The full pipeline of AlignNet is shown in Figure \ref{fig:pipeline}. It is designed based on the following principles: (1) spatial and temporal attention modules highlight the important regions; (2) learnable feature pyramids extract features from visual and audio inputs at multiple levels (temporal scales); (3) inspired by \cite{sun2018PWC-Net}, warping layers warp the lower-level reference modality feature based on the correspondence estimation on the current level to better predict distortion on larger scales; (4) correlation layers models the affinity between video features with audio features; (5) final dense correspondence estimation is predicted from the affinity map. We will explain each module in details in the following paragraphs.

\subsection{Spatial and Temporal Attention Modules}

Intuitively, certain keypoints are more reflective of the visual rhythms than others, so are certain sound fragments of the auditory beats. Hence, we use attention modules to highlight these important parts. Applying spatio-temporal attention jointly requires excessive parameters, so we propose to decouple it, using a spatial attention module and a temporal attention module.

For the spatial attention, we assign a weight to each keypoint, indicating the importance of this spatial location. To make the learning more effective and meaningful, we force the attention of symmetric keypoints (\textit{e.g.} left and right elbow) to be the same. 
Note that we assume keypoint attentions are the same over the whole dataset, so they are learned as model parameters (the number of parameters is the same as the number of keypoints).

For the temporal attention, we train a self-attention module conditioned on the input audio, and re-weight audio features at each time step accordingly. Different from keypoint attention, temporal attention varies with the input data, so the attention weights are the intermediate outputs of the model.

\subsection{Multimodal Feature Pyramid Extraction}
For each modality, we generate an $N$-level feature pyramid, with the 0-th level being the input features, \textit{i.e.}, $\mathbf{F_p^0}=\mathbf{F}_{v/a}^{t}, t\in(1,..,n/m)$. Convolutional layers then extract the feature representation at the $l$-th level, $\mathbf{F_p^l}$, from the feature representation at the ($l-1$)-th level, $\mathbf{F_p^{l-1}}$. 

\subsection{Time Warping}
The concept of warping is common to flow-based methods and goes back to~\cite{lucas1981iterative}. We introduce time warping layer to warp the video features given the estimated correspondences. The warping operation assesses previous errors and computes an incremental update, helping the prediction module to focus on the residual misalignment between video and audio. Formally, at the $l$-th level, we warp reference audio features toward video features using the upsampled prediction from the ($l+1$)-th level, $\mathbf{d}^{l+1}$:
\begin{equation}
    \mathbf{F}_{w}^{l}(\mathbf{x})=\mathbf{F}_{a}^{l}\left(\mathrm{up}_{\gamma}\left(\mathbf{d}^{l+1}\right)(\mathbf{x})\right)
\end{equation}
where $\mathbf{x}$ is the frame indices and $\mathrm{up}_{\gamma}$ is the upsampling function with scale factor $\gamma$. 

\subsection{Affinity Function}
In order to better model the correspondence, 
we employ an affinity function $A$ at each feature pyramid level. $A$ provides a measure of similarity between video features $\mathbf{F_v^l}^{t}, \: t\in(1,..,n^l)$ and audio feature $\mathbf{F_a^l}^{t}, \: t\in(1,..,m^l)$, denoted as $A\left(\mathbf{F_p}_{v}^{l}, \mathbf{F_p}_{a}^{l}\right)$.
The affinity is the dot product between embeddings: for $\mathbf{F_v^l}^{i}$ and $\mathbf{F_a^l}^{j}$,
\begin{equation}
    A(j, i)=\frac{\left({\mathbf{F_a^l}^{j}}^{\top} \mathbf{F_v^l}^{i}\right)}{\sum_{j} \left({\mathbf{F_a^l}^{j}}^{\top} \mathbf{F_v^l}^{i}\right)}
\end{equation}

\subsection{Multimodal Dense Correspondence Estimation}

Correspondences are normalized to $(-1,1)$. For a video with $N$ frames, we first include a correspondence loss, namely an L1 regression loss of the predicted correspondence at all pyramid levels:
\begin{equation}
    L_{fs}=\sum_{l=0}^{N}\lambda_{l} \sum_{i=0}^{n^l}\left|\mathbf{d}^{l}\left(i\right)-\tilde{\mathbf{d}}^{l}\left(i\right)\right|
\end{equation}
where $\mathbf{d}^{l}\left(i\right)$ is the ground truth correspondence and $\tilde{\mathbf{d}}^{l}\left(i\right)$ is the predicted correspondence of frame $i$ at level $l$, and $\lambda_{l}$ is the weighting factor of level $l$.

To force the network to predict realistic correspondence that are temporally monotonic, we further incorporate a monotonic loss:
\begin{equation}
    L_{mono}=\sum_{l=0}^{N}\lambda_{l}\sum_{i=0}^{n^{l}-2}\max\left(0,1 - \tilde{\mathbf{d}}^{l}\left(i\right) + \tilde{\mathbf{d}}^{l}\left(i+1\right)\right)
\end{equation}
With $\mu$ being a weighting hyperparameter, our full objective to minimize is therefore:
\begin{equation}
    loss=L_{fs}+\mu L_{mono}
\end{equation}

\subsection{Model Architecture Details}
AlignNet takes in normalized log-mel spectrograms with 128 frequency bins as audio input and pose/lip keypoints as visual input with shape (2$\times$\#keypoints, \#frames), where the 2 accounts for two dimensional (x,y) coordinates. Note that all convolutional layers in our network are 1D convolutions, which speed up the prediction at both training and testing time. We use a 4-level feature pyramid, with channels 128, 64, 32, 16 and the temporal downscale factors 1/3, 1/2, 1/2, 1/2, respectively. The final output is obtained by upsampling the 1st-level correspondence to match with the input dimension. Note that our model can take in videos of arbitrary lengths.
\section{Experiments}

We evaluate AlignNet on two tasks: dance-music alignment, and lip-speech alignment. We test our method, both quantitatively and qualitatively on the these tasks, and compare our results with two benchmark methods. Finally, subjective user studies are performed for dance and speech retargeting.
\vspace{-0.5em}
\paragraph{Baselines} In order to compare our methods with other state-of-the-art audio-video synchronization techniques, we implemented two baselines: SyncNet~\cite{chung2016out} and SyncNet+DTW ~\cite{halperin2019dynamic}. For SyncNet, we follow the exact steps described in \cite{chung2016out} except that we replace raw frame inputs with pose/lip keypoints and replace MFCC feature inputs with normalized log-mel spectrograms. Since (1) the relevant features to dancing and speaking are essentially body and lip motion and (2) MFCC features are handcrafted and low-dimensional (\textit{i.e.} contain less information than spectrograms), these changes should reach similar performance of the original method, if not better. SyncNet+DTW uses the pre-trained SyncNet as their feature extractor, so we also replace the SyncNet here with our implementation. Then, DTW is applied on the extracted video and audio features.
\vspace{-0.5em}
\paragraph{Training}
During training, we use online sample generation for both our method and the baselines. In our method, the video features are randomly distorted and scaled temporally each time to prevent the network from overfitting finite misalignment patterns. To better mimic real world situations, the distortions are always temporally monotonic, and we linearly interpolate the keypoints from adjacent frames. For the baselines, SyncNet is trained with random pairs of matching / non-matching pairs with the same clip length and maximum temporal difference as the original paper~\cite{chung2016out}. We further augment the training data by horizontally flipping all poses in a video clip, which is a natural choice because visual dynamics is mirror-invariant. For the audio inputs, we employ the time masking and frequency masking suggested by SpecAugment~\cite{park2019SpecAugmentAS}. \textit{Adam} is used to optimize the network parameters with a learning rate of $3\times10^{-4}$.
\vspace{-0.5em}
\paragraph{Evaluation} 
Quantitative evaluation was performed using a human perception-inspired metric, based on the maximum acceptable audio-visual asynchrony used in the broadcasting industry. According to the International Telecommunications Union (ITU), the auditory signal should not lag by more than 125 ms or lead by more than 45 ms. Therefore, the accuracy metric we use is the percentage of frames in the aligned signal which fall inside of the above acceptable (human undetectable) range, compared to the ground truth alignment.

We also defined Average Frame Error (AFE) as the average difference between the reconstructed frame indices and the original undistorted frame indices. This gives a more direct measure of how close the reconstructed video is compared to the original video.

\subsection{Dance-music Alignment}
\label{section:dance_align}

\paragraph{Dance50 Dataset}
We introduce \textit{Dance50} dataset for dance-music alignment. The dataset contains 50 hours of dancing clips from over 200 dancers. There are around 10,000 12-second videos in the training set, 2,000 in the validation set, and 2,000 in the testing set. Our training, validation, and testing sets contain disjoint videos, such that there is no overlap between the videos from any two sets. All dancing clips are collected from K-pop dance cover videos from \textit{YouTube.com} and \textit{Bilibili.com}. Similar to \cite{ginosar2019gestures}, we represent our annotations of the poses with a temporal sequence of 2D skeleton keypoints, obtained using OpenPose~\cite{cao2018openpose} BODY\_25 model, discarding 6 noisy keypoints from the feet and keeping the rest.
We refer the readers to supplementary material for more details.

\paragraph{Settings}
We use \textit{Dance50} to evaluate the performance of method on dance synchronization. We compare our method against the state-of-the-art baselines using the aforementioned evaluation metrics. 
\vspace{-0.5em}
\paragraph{Results}
\begin{table}
\begin{center}
\begin{tabular}{c|c|c}
\hline
\multirow{2}{*}{Methods} & \multicolumn{2}{c}{Performance}\\ \cline{2-3}
 & AFE & Accuracy\\ \hline
SyncNet~\cite{chung2016out} & 6.58 & 25.88\\ \hline
SyncNet+DTW~\cite{halperin2019dynamic} & 4.27 & 38.29\\\hline
AlignNet (Ours)& \textbf{0.94} & \textbf{89.60}\\ \hline
\end{tabular}
\caption{Performance of music-dance alignment on \textit{Dance50} dataset.}
\label{table:dance}
\end{center}
\vspace{-2em}
\end{table}

Table~\ref{table:dance} compares the performance of our method and the two baselines on \textit{Dance50} testing set. Our method significantly reduces the AFE, and achieves a gain of 51.31\% in accuracy, which reaches 89.60\%. Our network obtains a satisfactory performance even on video clips completely unseen in training, and this shows that the proposed method effectively learns the implicit mapping between video dynamics and audio features on multiple temporal scales. 

\begin{figure*}
\begin{center}
  \includegraphics[width=\textwidth]{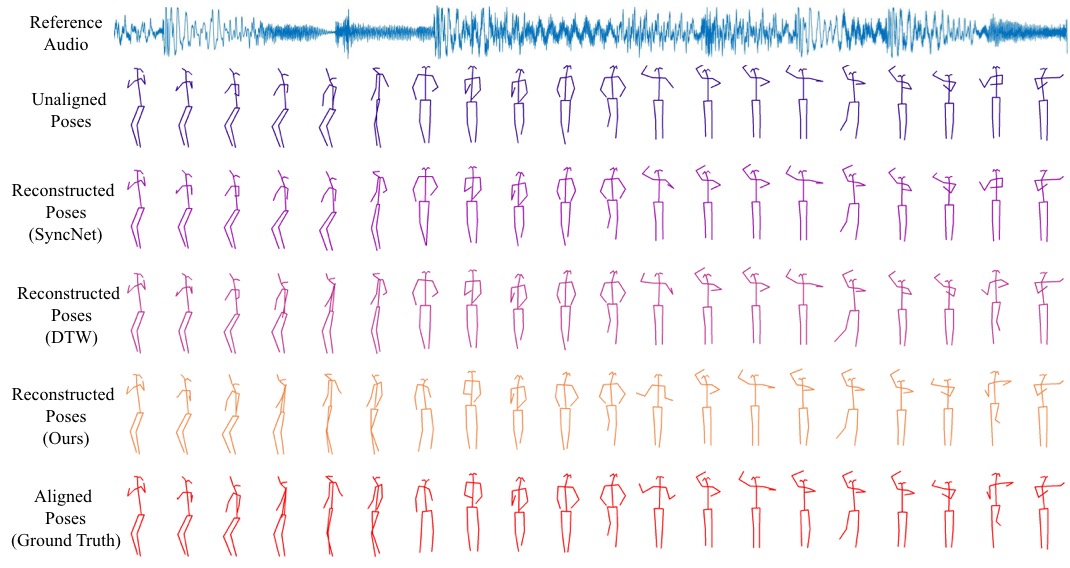}
  \caption{Skeleton visualizations of the proposed method and comparison with the baselines on \textit{Dance50}. For better visualization, we evenly sample 20 frames from the original 180 frames.}
  \label{fig:dancedemo}   
\end{center}
\vspace{-1em}
\end{figure*}

To get a better idea of how well our method aligns the video to the audio compared to the baselines methods, we show some visualization of the skeletons reconstructed from the correspondence predictions in Figure \ref{fig:dancedemo}. It can be seen that our method closely recovers the original aligned poses, while both SyncNet and DTW fails to align poses warped on different temporal scales. 

\begin{figure}[!ht]
    \centering
    \includegraphics[width=0.48\textwidth]{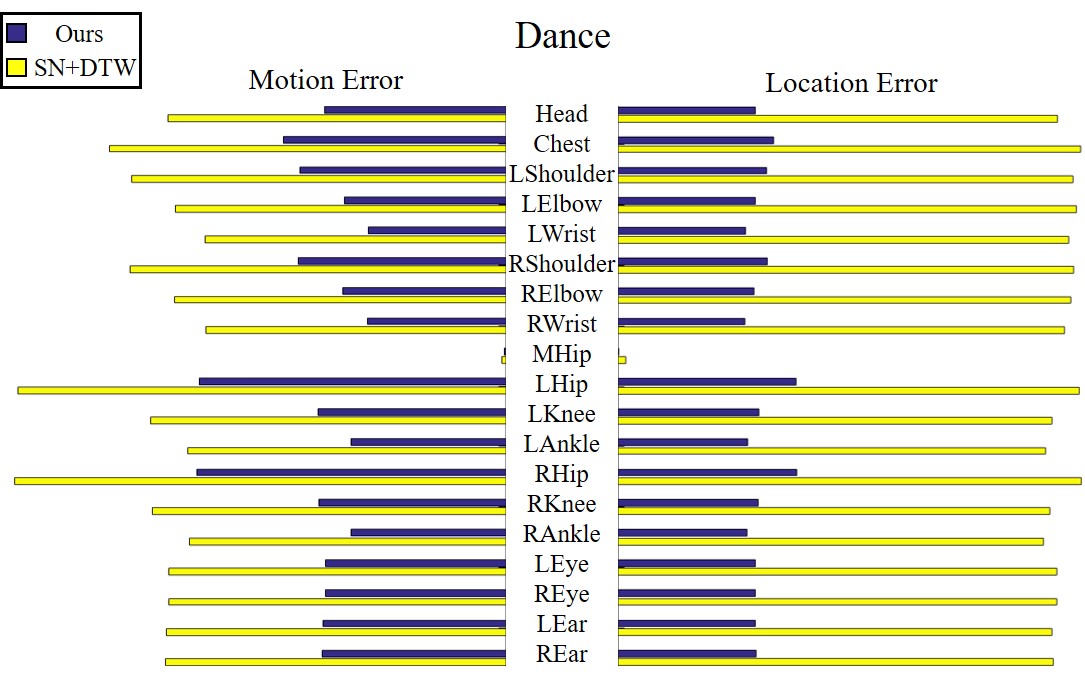}
    \caption{Comparison of the performance (mean motion \& location error) of our method and SyncNet + DTW (SN+DTW) on the dance-music alignment task.}
    \label{fig:pose_error}
\end{figure}

We also show the mean motion error and location error of each human keypoint after alignment in Figure~\ref{fig:pose_error}. The motion error reflects visual rhythm difference between aligned video and original video (Figure~\ref{fig:pose_error} Left), while the location error reflects the objective difference between aligned video and original video (Figure~\ref{fig:pose_error} Right). For motion error, we normalize the keypoint velocity in pixel space of both x and y coordinates to (-1,1). Similarly, we normalize the location of each keypoint for calculating location error. As shown in the figure, for both motion and location, our proposed method outperforms the baseline method consistently by a large margin under all human keypoints. It's worth noticing that mid-hip locations are always subtracted during preprocessing. Thus, the mid-hip error of both motion and location is significantly smaller than other keypoints. 

Furthermore, we plot examples of the ground truth distortion compared to the distortion predicted by SyncNet-DTW and our method in Figure \ref{fig:alignmentcurve} (left). It can be seen that SyncNet-DTW captures some of the mapping between video and audio very coarsely, while failing to align with more subtle changes. We believe that this is because SyncNet-DTW uses SyncNet as its backbone network, which is trained with blocks of matching / non-matching pairs, assuming that the frames within each block is temporally aligned. In our application scenario, however, the video could be distorted at any moment on any temporal scale, so their method cannot deal with such distortions very well. In contrast, the prediction from our method closely matches the ground truth distortion and is able to undistort \& recover the input video. We believe that this is because (1) the hierarchical nature of our model helps the network extract visual and auditory features on multiple temporal scales; (2) the warping layer makes it easier to predict coarser trends on larger temporal scales; and (3) the affinity map helps to correlate the relationship between video features and audio features. We also show the effectiveness of each model in the ablation study later. 

\subsection{Lip-speech Alignment}

\paragraph{Settings}

To evaluate our model on the lip-speech alignment task, we assemble a subset from the \textit{VoxCeleb2} dataset~\cite{Chung2018VoxCeleb2}, which is an audio-visual dataset consisting of short clips of human speech extracted from YouTube videos. We select and cut 22000 video clips of length 12 seconds and with 25 fps, mounting to 73 hours in total. The dataset contains 333 different identities and a wide range of ethnicity and languages. The dataset split is as follows: the training set has 18000 clips, and the validation and testing sets each have 2000 clips. The sets are exclusively video-disjoint, and there is no overlap between any pair of videos from any two sets. Similar to \textit{Dance50}, to deal with the noisy frame-wise results, we first remove the outlier keypoints by performing median filtering and then the missing values are linearly interpolated and smoothed with a Savitzky-Golay filter. Again, we compare our method against the state-of-the-art baselines using the evaluation metrics mentioned above, and all methods are trained with online sample generation and data augmentation, for 500 epochs, and the best performing model on the validation set is selected for testing.

\paragraph{Results}

\begin{figure}
    \centering
    \includegraphics[width=0.48\textwidth]{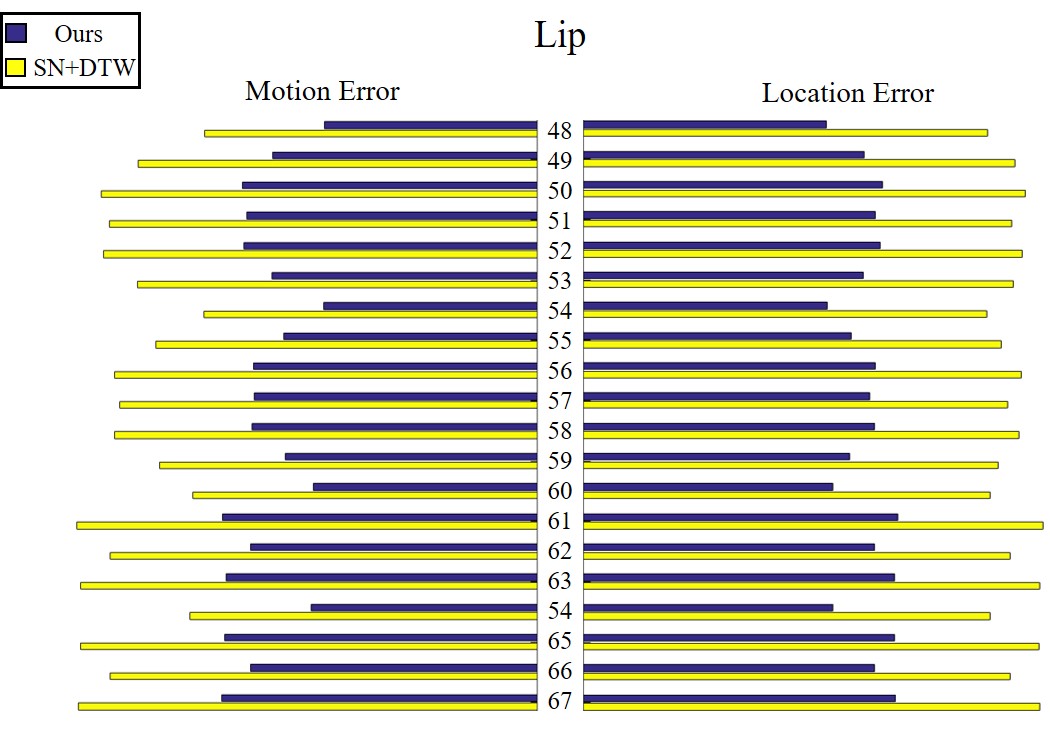}
    \caption{Comparison of the performance (mean motion \& location error) of our method and SyncNet + DTW (SN+DTW) on the lip-speech alignment task.}
    \label{fig:lip_error}
\end{figure}

Similar to Section~\ref{section:dance_align}, we show the mean motion error and location error of each lip keypoint after alignment in Figure \ref{fig:lip_error}. Keypoints labels are the same as used in OpenPose~\cite{cao2018openpose}. As shown in the figure, for both motion and location, our proposed method outperforms the baseline method consistently by a large margin for all the lip keypoints.

\begin{table}
\begin{center}
\begin{tabular}{c|c|c}
\hline
\multirow{2}{*}{Methods} & \multicolumn{2}{c}{Performance}\\ \cline{2-3}
 & AFE & Accuracy\\ \hline
SyncNet~\cite{chung2016out} & 6.33 & 27.41 \\ \hline
SyncNet+DTW~\cite{halperin2019dynamic} & 3.96 & 35.94\\ \hline
AlignNet\:(Ours) & \textbf{1.03} & \textbf{81.05}\\ \hline
\end{tabular}
\end{center}
\caption{Performance of speech-lip alignment on \textit{VoxCeleb2} dataset.}
\label{table:lip} 
\end{table}

Table~\ref{table:lip} compares the performance of our method and the two baselines on subset of \textit{VoxCeleb2} testing set. Our method significantly reduces the AFE , and achieves a gain of 45.11\% in accuracy, which achieves 81.05\%. Note that here the AFE is similar to the AFE on \textit{Dance50}, but the accuracy on \textit{Dance50} is much higher. This is because the frame rates of the two datasets are different (30 fps for \textit{Dance50} and 25 fps for \textit{VoxCeleb2}), and one frame in \textit{VoxCeleb2} corresponds to a longer duration in ms. Therefore, these results do not suggest the relative difficulty of dance-music alignment and speech-lip alignment, but they together indicate that our proposed method for audio-video alignment works well in different application scenarios.

\begin{figure}
    \centering
    \includegraphics[width=0.48\textwidth]{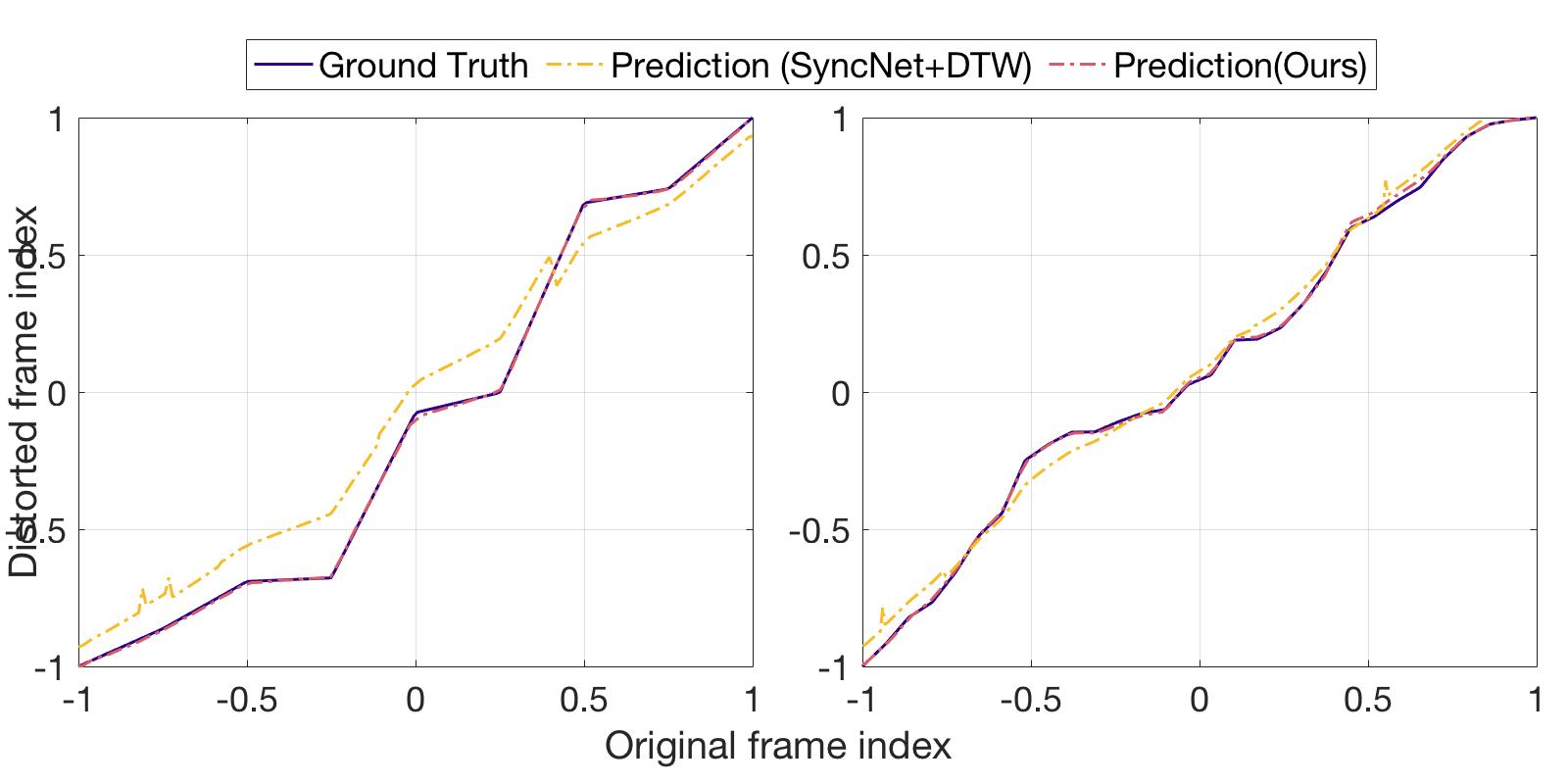}
    \caption{Qualitative results of the ground truth distortion compared to the predicted distortions by SyncNet-DTW and our method. Left: dance-music alignment, Right: lip-speech alignment.}
    \label{fig:alignmentcurve}
\end{figure}

Similar to the dance-music alignment task, we plot an example of the ground truth distortion compared to the distortion predicted by DTW and our method in the bottom part of Figure \ref{fig:alignmentcurve} (Right). This result clearly shows that our method recover the distorted video at a very high accuracy. 

\subsection{Ablation Study}
To show the effectiveness of each module in our method, we conduct an ablation study on the \textit{Dance50} dataset, where we add the modules sequentially to the pipeline and evaluate the performance. The training setup is the same as before. We start with a basic network, denoted as base, where direct pose keypoints are fed in as inputs and no feature pyramid is used, predicting the correspondence directly from an affinity map. Then, the following modules are added sequentially to the network: FP (feature pyramid), MI (motion inputs, velocity and acceleration), SA (spectrogram augmentation from \cite{park2019SpecAugmentAS}), KA (keypoint attention) and TA (temporal attention). Table \ref{table:ablation} shows the results of the ablation experiments. It is clear that both feature pyramid and motion inputs boost the performance of the network by a large margin, and spectrogram augmentation further helps the network to learn meaningful feature even when some temporal and frequency information are missing and helps alleviate overfitting.

\begin{table}
\begin{center}
\begin{tabular}{c|c|c}
\hline
\multirow{2}{*}{Methods} & \multicolumn{2}{c}{Performance}\\ \cline{2-3}
 & AFE & Accuracy\\ \hline
Base & 2.87 & 43.57 \\ \hline
FP & 2.45 & 56.88\\ \hline
FP+MI & 1.81 & 67.49\\ \hline
FP+MI+SA & 1.32 & 75.33\\ \hline
FP+MI+SA+TA & 1.14 & 79.65\\ \hline
FP+MI+SA+KA & 0.97 & 88.20\\ \hline
FP+MI+SA+KA+TA & \textbf{0.94} & \textbf{89.60}\\ \hline
\end{tabular}
\end{center}
\caption{Ablation study on dance-music alignment by adding these modules sequentially: Base (base model), FP (feature pyramid), MI (motion inputs, velocity and acceleration), SA (spectrogram augmentation), KA (keypoint attention) and TA (temporal attention).}
\label{table:ablation} 
\end{table}

Both keypoints attention and temporal attention can improve the alignment performance. This means some keypoints and time steps are more important for alignment. And focusing on these spatial temporal regions will significantly improve the performance. However, temporal attention is not as effective as keypoints attention. One possible reason is that most music pieces have fixed and similar beats, which can be implicitly encoded in the network without temporal attention module. Thus, adding temporal might have limited improvement. On the other hand, motion rhythms are not always as clear as music beats, \textit{i.e.} dancer might dance differently given the same music piece. Thus, focusing on specific keypoints can significantly improve the performance of our proposed module.  

\subsection{Attention Visualizations}

\begin{figure}
    \centering
    \includegraphics[width=0.4\textwidth]{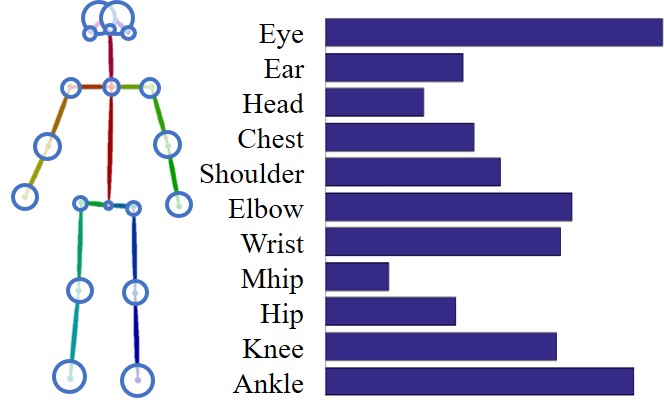}
    \caption{Keypoint attention visualization for the dance-music alignment task. Circle sizes and bar lengths represent the attention magnitudes on human keypoints.}
    \label{fig:pose_attention}
\end{figure}

\begin{figure}
    \centering
    \includegraphics[width=0.48\textwidth]{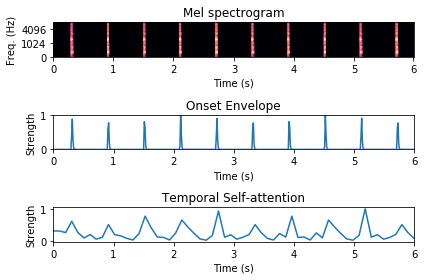}
    \caption{Mel-spectrogram (top), onset envelop (middle), and temporal attention (bottom) on a sample audio. Attention magnitudes agree with audio onsets.}
    \label{fig:temporal_attention}
\end{figure}

\paragraph{Keypoint Attention}
We visualize the learned keypoint attentions by our model for the dance-music alignment task in Figure \ref{fig:pose_attention}, where a skeleton visualization is presented alongside the histogram of the attention values after softmax. A circle is drawn around each keypoint, and the radii are larger for keypoints with larger attention values. Results show the attention for eyes and limbs (elbows, wrists, knees, ankles) are significantly larger than the attention for the bulk of the body (chest, shoulders, hip). The network has learnt to rely more on the limbs whose movements are more rapid, and less on the bulk parts that seldom moved very fast. We also note that the network assigns more attention to the eyes but very little to head or ears. A possible explanation for this is that the movements of head (like turns) is best represented by tracking two eyes, rather than ears that are sometimes occluded.
\vspace{-1em}
\paragraph{Temporal Attention}
Figure~\ref{fig:temporal_attention} shows the temporal attention on an example audio mel-spectrogram. As can be seen, the attention module tends to focus more on the onsets of a music piece which can be easily understood since dancers always switch motion pattern at these onsets. Thus, focusing more on these temporal regions can help with better alignment.

\subsection{Dance and Speech Retargeting}

\begin{table}
\begin{center}
\begin{tabular}{c|cc}
\hline
\multirow{2}{*}{Task} & \multicolumn{2}{c}{Percentage}\\ \cline{2-3}
 & Direct Combination & Ours\\ \hline
Dance Retargeting & 31\% & 69\% \\ \hline
Speech Retargeting & 22\% & 78\% \\ \hline
\end{tabular}
\caption{Subjective user study for dance and speech retargeting.}
\label{table:subjective}
\end{center}
\vspace{-1em}
\end{table}

By time-warping the visual beats of existing dance footage into alignment with new music, we can change the song that a performer is dancing to, which is known as dancing retargeting. We generate 7 evaluation pairs for dance retargeting evaluation. Each pair contains a dancing video directly combined with another music piece and the same video retargeted according to the audio using AlignNet. We ask 13 people for their preference of all the evaluation pairs, and the result is shown in Table~\ref{table:subjective}. Similarly, we generate 7 evaluation pairs for speech retargeting evaluation. It's worth noticing that in these speech pairs, both videos say the same sentence since it is not reasonable to align talking face with different sentences. As can be seen in Table~\ref{table:subjective}, subjects prefer ours to direct combinations, which means that our method can generalize well to real-world data. However, our method performs slightly worse in dance retargeting. The main reason is most dancing musics have similar beats and direct combination can achieve decent performance. Demo videos of both the retargeting task and the synthetic re-alignment task can be found in the supplementary materials and on our webpage.
\section{Conclusion}

In conclusion, we proposed AlignNet, an end-to-end model to address the problem of audio-video alignment at arbitrary temporal scale. The approach adopts the following principles: attention, pyramidal processing, warping, and affinity function. AlignNet establishes new state-of-the-art performance on dance-music alignment and speech-lip alignment.
We hope our work can inspire more studies in video and audio synchronization under non-uniform and irregular misalignment (\textit{e.g.} dancing).  
\section*{Appendix 1: Dance50 Dataset}

We introduce \textit{Dance50} dataset for dance-music alignment. The dataset contains 50 hours of dancing clips from over 200 dancers. There are 20,000 6-second videos in the training set, 5,000 in the validation set, and 5,000 in the testing set. Our training, validation, and testing sets contain disjoint videos, such that there is no overlap between the videos from any two sets.

\paragraph{Dataset collection}
In the construction of the \textit{Dance50} dataset, we focus on dancing videos in which the visual dynamics and audio beats are closely coupled and well synchronized. Therefore, we crawled K-pop dance cover videos performed by experienced dancers from \textit{YouTube} and \textit{Bilibili}. All videos have a video frame rate of 30 fps and an audio sampling rate of 44,100 Hz. The videos are then cut into 6-second clips (180 frames per clip). We further filter the video clips to include only high-quality samples satisfying the following requirements: (1) single-person, (2) minimal or no camera movement, (3) full size shots with minimal or no occlusion of body parts. These restrictions avoids the problem of person tracking / ReID and keeps the dataset clean from excessive external noise.

\paragraph{Dataset annotations}
Similar to \cite{ginosar2019gestures}, we represent our annotations of the dancers' pose over time with a temporal sequence of 2D skeleton keypoints, obtained using OpenPose~\cite{cao2018openpose} BODY\_25 model. We then discard 3 very noisy keypoints from each foot, keeping 19 keypoints to represent poses. To cope with the noisy framewise results, we first remove the outlier keypoints by performing median filtering, and then the missing values are linearly interpolated and smoothed with a Savitzky-Golay filter. We provide these annotations with the videos at 30 fps. Furthermore, the quality of these annotations are checked by comparing detected keypoints with manually labeled keypoints on a subset of the training data.

\section*{Appendix 2:More Experiments}

\subsection*{Pose Attention}

\begin{table}[H]
\begin{tabular}{l|c|c|c|c}
\hline
\multirow{3}{*}{Keypoint} & \multicolumn{4}{c}{L1 error (normalized)} \\ \cline{2-5} 
 & \multicolumn{2}{c}{Location} & \multicolumn{2}{c}{Velocity} \\ \cline{2-5} 
 & Ours & SN+DTW & Ours & SN+DTW \\ \hline
Head & 0.128 & 0.412 & 0.112 & 0.209 \\ \hline
Chest & 0.146 & 0.433 & 0.137 & 0.246 \\ \hline
Left Shoulder & 0.139 & 0.426 & 0.127 & 0.232 \\ \hline
Left Elbow & 0.128 & 0.429 & 0.099 & 0.205 \\ \hline
Left Wrist & 0.119 & 0.422 & 0.085 & 0.186 \\ \hline
Right Shoulder & 0.139 & 0.427 & 0.128 & 0.233 \\ \hline
Right Elbow & 0.127 & 0.424 & 0.101 & 0.205 \\ \hline
Right Wrist & 0.118 & 0.418 & 0.086 & 0.186 \\ \hline
Mid Hip & 0 & 0 & 0 & 0 \\ \hline
Left Hip & 0.166 & 0.432 & 0.189 & 0.302 \\ \hline
Left Knee & 0.131 & 0.406 & 0.116 & 0.220 \\ \hline
Left Ankle & 0.121 & 0.400 & 0.096 & 0.197 \\ \hline
Right Hip & 0.167 & 0.434 & 0.191 & 0.305 \\ \hline
Right Knee & 0.131 & 0.405 & 0.116 & 0.219 \\ \hline
Right Ankle & 0.121 & 0.398 & 0.096 & 0.196 \\ \hline
Left Eye & 0.128 & 0.411 & 0.112 & 0.209 \\ \hline
Right Eye & 0.128 & 0.411 & 0.112 & 0.209 \\ \hline
Left Ear & 0.128 & 0.407 & 0.113 & 0.210 \\ \hline
Right Ear & 0.129 & 0.408 & 0.113 & 0.211 \\ \hline
\end{tabular}
\caption{Pose Attention.}
\end{table}

\subsection*{Lip Attention}

\begin{figure}[H]
    \centering
    \includegraphics[width=0.48\textwidth]{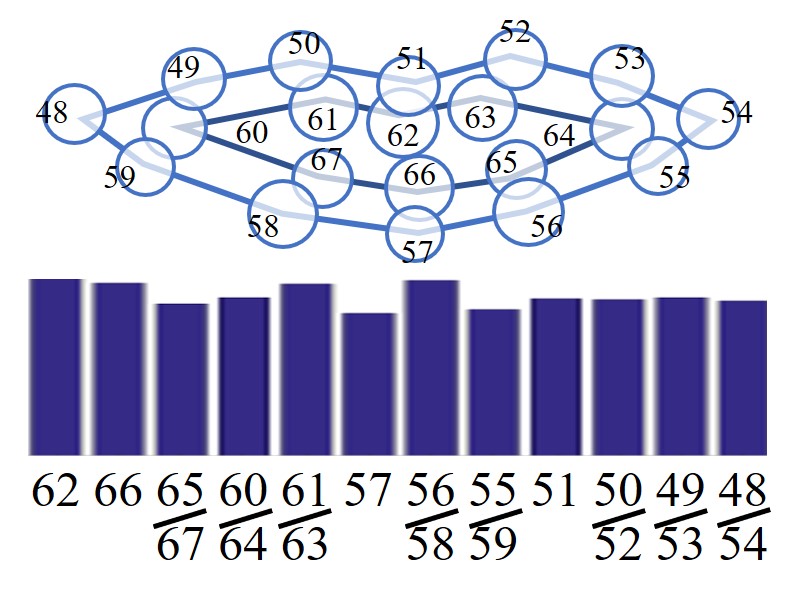}
    \caption{Lip attention Visualization}
    \label{fig:lip_attention}
\end{figure}

\begin{table}[H]
\begin{tabular}{c|c|c|c|c}
\hline
\multirow{3}{*}{Keypoint} & \multicolumn{4}{c}{L1 error (normalized)} \\ \cline{2-5} 
 & \multicolumn{2}{c}{Location} & \multicolumn{2}{c}{Velocity} \\ \cline{2-5} 
 & Ours & SN+DTW & Ours & SN+DTW \\ \hline
48 & 0.265 & 0.441 & 0.188 & 0.295 \\ \hline
49 & 0.306 & 0.471 & 0.235 & 0.355 \\ \hline
50 & 0.326 & 0.482 & 0.262 & 0.388 \\ \hline
51 & 0.318 & 0.468 & 0.258 & 0.380 \\ \hline
52 & 0.324 & 0.479 & 0.260 & 0.386 \\ \hline
53 & 0.305 & 0.470 & 0.236 & 0.355 \\ \hline
54 & 0.266 & 0.441 & 0.189 & 0.296 \\ \hline
55 & 0.292 & 0.456 & 0.225 & 0.339 \\ \hline
56 & 0.318 & 0.478 & 0.252 & 0.376 \\ \hline
57 & 0.312 & 0.463 & 0.251 & 0.371 \\ \hline
58 & 0.318 & 0.475 & 0.253 & 0.375 \\ \hline
59 & 0.290 & 0.453 & 0.224 & 0.335 \\ \hline
60 & 0.272 & 0.444 & 0.199 & 0.306 \\ \hline
61 & 0.343 & 0.502 & 0.280 & 0.409 \\ \hline
62 & 0.318 & 0.466 & 0.255 & 0.379 \\ \hline
63 & 0.340 & 0.498 & 0.276 & 0.405 \\ \hline
64 & 0.272 & 0.444 & 0.200 & 0.308 \\ \hline
65 & 0.339 & 0.497 & 0.278 & 0.406 \\ \hline
66 & 0.318 & 0.466 & 0.255 & 0.379 \\ \hline
67 & 0.340 & 0.498 & 0.280 & 0.408 \\ \hline
\end{tabular}
\caption{Lip Attention}
\end{table}

{\small
\bibliographystyle{ieee}
\bibliography{egbib}
}

\end{document}